\documentclass{article}
\usepackage{arxiv}

\usepackage[utf8]{inputenc} 
\usepackage[T1]{fontenc}    
\usepackage{hyperref}       
\usepackage{url}            
\usepackage{booktabs}       
\usepackage{amsmath}
\usepackage{amsfonts}       
\usepackage{nicefrac}       
\usepackage{microtype}      
\usepackage{algorithm2e}
\usepackage{bbm}
\usepackage[table,xcdraw]{xcolor}
\usepackage{graphicx}
\usepackage{subcaption}
\usepackage{bbding}
\usepackage{float}
\usepackage{grffile}
\usepackage[draft]{minted}  
\newminted{python}{fontsize=\small}

\makeatletter
\def\PYGdefault@reset{\let\PYGdefault@it=\relax \let\PYGdefault@bf=\relax%
    \let\PYGdefault@ul=\relax \let\PYGdefault@tc=\relax%
    \let\PYGdefault@bc=\relax \let\PYGdefault@ff=\relax}
\def\PYGdefault@tok#1{\csname PYGdefault@tok@#1\endcsname}
\def\PYGdefault@toks#1+{\ifx\relax#1\empty\else%
    \PYGdefault@tok{#1}\expandafter\PYGdefault@toks\fi}
\def\PYGdefault@do#1{\PYGdefault@bc{\PYGdefault@tc{\PYGdefault@ul{%
    \PYGdefault@it{\PYGdefault@bf{\PYGdefault@ff{#1}}}}}}}
\def\PYGdefault#1#2{\PYGdefault@reset\PYGdefault@toks#1+\relax+\PYGdefault@do{#2}}

\expandafter\def\csname PYGdefault@tok@kn\endcsname{\let\PYGdefault@bf=\textbf\def\PYGdefault@tc##1{\textcolor[rgb]{0.00,0.50,0.00}{##1}}}
\expandafter\def\csname PYGdefault@tok@ni\endcsname{\let\PYGdefault@bf=\textbf\def\PYGdefault@tc##1{\textcolor[rgb]{0.60,0.60,0.60}{##1}}}
\expandafter\def\csname PYGdefault@tok@nb\endcsname{\def\PYGdefault@tc##1{\textcolor[rgb]{0.00,0.50,0.00}{##1}}}
\expandafter\def\csname PYGdefault@tok@ch\endcsname{\let\PYGdefault@it=\textit\def\PYGdefault@tc##1{\textcolor[rgb]{0.25,0.50,0.50}{##1}}}
\expandafter\def\csname PYGdefault@tok@ge\endcsname{\let\PYGdefault@it=\textit}
\expandafter\def\csname PYGdefault@tok@dl\endcsname{\def\PYGdefault@tc##1{\textcolor[rgb]{0.73,0.13,0.13}{##1}}}
\expandafter\def\csname PYGdefault@tok@s1\endcsname{\def\PYGdefault@tc##1{\textcolor[rgb]{0.73,0.13,0.13}{##1}}}
\expandafter\def\csname PYGdefault@tok@sd\endcsname{\let\PYGdefault@it=\textit\def\PYGdefault@tc##1{\textcolor[rgb]{0.73,0.13,0.13}{##1}}}
\expandafter\def\csname PYGdefault@tok@vg\endcsname{\def\PYGdefault@tc##1{\textcolor[rgb]{0.10,0.09,0.49}{##1}}}
\expandafter\def\csname PYGdefault@tok@nf\endcsname{\def\PYGdefault@tc##1{\textcolor[rgb]{0.00,0.00,1.00}{##1}}}
\expandafter\def\csname PYGdefault@tok@gs\endcsname{\let\PYGdefault@bf=\textbf}
\expandafter\def\csname PYGdefault@tok@o\endcsname{\def\PYGdefault@tc##1{\textcolor[rgb]{0.40,0.40,0.40}{##1}}}
\expandafter\def\csname PYGdefault@tok@nc\endcsname{\let\PYGdefault@bf=\textbf\def\PYGdefault@tc##1{\textcolor[rgb]{0.00,0.00,1.00}{##1}}}
\expandafter\def\csname PYGdefault@tok@err\endcsname{\def\PYGdefault@bc##1{\setlength{\fboxsep}{0pt}\fcolorbox[rgb]{1.00,0.00,0.00}{1,1,1}{\strut ##1}}}
\expandafter\def\csname PYGdefault@tok@il\endcsname{\def\PYGdefault@tc##1{\textcolor[rgb]{0.40,0.40,0.40}{##1}}}
\expandafter\def\csname PYGdefault@tok@gt\endcsname{\def\PYGdefault@tc##1{\textcolor[rgb]{0.00,0.27,0.87}{##1}}}
\expandafter\def\csname PYGdefault@tok@gp\endcsname{\let\PYGdefault@bf=\textbf\def\PYGdefault@tc##1{\textcolor[rgb]{0.00,0.00,0.50}{##1}}}
\expandafter\def\csname PYGdefault@tok@cm\endcsname{\let\PYGdefault@it=\textit\def\PYGdefault@tc##1{\textcolor[rgb]{0.25,0.50,0.50}{##1}}}
\expandafter\def\csname PYGdefault@tok@si\endcsname{\let\PYGdefault@bf=\textbf\def\PYGdefault@tc##1{\textcolor[rgb]{0.73,0.40,0.53}{##1}}}
\expandafter\def\csname PYGdefault@tok@sa\endcsname{\def\PYGdefault@tc##1{\textcolor[rgb]{0.73,0.13,0.13}{##1}}}
\expandafter\def\csname PYGdefault@tok@sr\endcsname{\def\PYGdefault@tc##1{\textcolor[rgb]{0.73,0.40,0.53}{##1}}}
\expandafter\def\csname PYGdefault@tok@sb\endcsname{\def\PYGdefault@tc##1{\textcolor[rgb]{0.73,0.13,0.13}{##1}}}
\expandafter\def\csname PYGdefault@tok@mb\endcsname{\def\PYGdefault@tc##1{\textcolor[rgb]{0.40,0.40,0.40}{##1}}}
\expandafter\def\csname PYGdefault@tok@c\endcsname{\let\PYGdefault@it=\textit\def\PYGdefault@tc##1{\textcolor[rgb]{0.25,0.50,0.50}{##1}}}
\expandafter\def\csname PYGdefault@tok@se\endcsname{\let\PYGdefault@bf=\textbf\def\PYGdefault@tc##1{\textcolor[rgb]{0.73,0.40,0.13}{##1}}}
\expandafter\def\csname PYGdefault@tok@mi\endcsname{\def\PYGdefault@tc##1{\textcolor[rgb]{0.40,0.40,0.40}{##1}}}
\expandafter\def\csname PYGdefault@tok@sc\endcsname{\def\PYGdefault@tc##1{\textcolor[rgb]{0.73,0.13,0.13}{##1}}}
\expandafter\def\csname PYGdefault@tok@gi\endcsname{\def\PYGdefault@tc##1{\textcolor[rgb]{0.00,0.63,0.00}{##1}}}
\expandafter\def\csname PYGdefault@tok@vi\endcsname{\def\PYGdefault@tc##1{\textcolor[rgb]{0.10,0.09,0.49}{##1}}}
\expandafter\def\csname PYGdefault@tok@gu\endcsname{\let\PYGdefault@bf=\textbf\def\PYGdefault@tc##1{\textcolor[rgb]{0.50,0.00,0.50}{##1}}}
\expandafter\def\csname PYGdefault@tok@kr\endcsname{\let\PYGdefault@bf=\textbf\def\PYGdefault@tc##1{\textcolor[rgb]{0.00,0.50,0.00}{##1}}}
\expandafter\def\csname PYGdefault@tok@kc\endcsname{\let\PYGdefault@bf=\textbf\def\PYGdefault@tc##1{\textcolor[rgb]{0.00,0.50,0.00}{##1}}}
\expandafter\def\csname PYGdefault@tok@kp\endcsname{\def\PYGdefault@tc##1{\textcolor[rgb]{0.00,0.50,0.00}{##1}}}
\expandafter\def\csname PYGdefault@tok@no\endcsname{\def\PYGdefault@tc##1{\textcolor[rgb]{0.53,0.00,0.00}{##1}}}
\expandafter\def\csname PYGdefault@tok@go\endcsname{\def\PYGdefault@tc##1{\textcolor[rgb]{0.53,0.53,0.53}{##1}}}
\expandafter\def\csname PYGdefault@tok@nt\endcsname{\let\PYGdefault@bf=\textbf\def\PYGdefault@tc##1{\textcolor[rgb]{0.00,0.50,0.00}{##1}}}
\expandafter\def\csname PYGdefault@tok@m\endcsname{\def\PYGdefault@tc##1{\textcolor[rgb]{0.40,0.40,0.40}{##1}}}
\expandafter\def\csname PYGdefault@tok@bp\endcsname{\def\PYGdefault@tc##1{\textcolor[rgb]{0.00,0.50,0.00}{##1}}}
\expandafter\def\csname PYGdefault@tok@nl\endcsname{\def\PYGdefault@tc##1{\textcolor[rgb]{0.63,0.63,0.00}{##1}}}
\expandafter\def\csname PYGdefault@tok@nv\endcsname{\def\PYGdefault@tc##1{\textcolor[rgb]{0.10,0.09,0.49}{##1}}}
\expandafter\def\csname PYGdefault@tok@cs\endcsname{\let\PYGdefault@it=\textit\def\PYGdefault@tc##1{\textcolor[rgb]{0.25,0.50,0.50}{##1}}}
\expandafter\def\csname PYGdefault@tok@gd\endcsname{\def\PYGdefault@tc##1{\textcolor[rgb]{0.63,0.00,0.00}{##1}}}
\expandafter\def\csname PYGdefault@tok@cpf\endcsname{\let\PYGdefault@it=\textit\def\PYGdefault@tc##1{\textcolor[rgb]{0.25,0.50,0.50}{##1}}}
\expandafter\def\csname PYGdefault@tok@sh\endcsname{\def\PYGdefault@tc##1{\textcolor[rgb]{0.73,0.13,0.13}{##1}}}
\expandafter\def\csname PYGdefault@tok@ow\endcsname{\let\PYGdefault@bf=\textbf\def\PYGdefault@tc##1{\textcolor[rgb]{0.67,0.13,1.00}{##1}}}
\expandafter\def\csname PYGdefault@tok@vm\endcsname{\def\PYGdefault@tc##1{\textcolor[rgb]{0.10,0.09,0.49}{##1}}}
\expandafter\def\csname PYGdefault@tok@kt\endcsname{\def\PYGdefault@tc##1{\textcolor[rgb]{0.69,0.00,0.25}{##1}}}
\expandafter\def\csname PYGdefault@tok@ss\endcsname{\def\PYGdefault@tc##1{\textcolor[rgb]{0.10,0.09,0.49}{##1}}}
\expandafter\def\csname PYGdefault@tok@k\endcsname{\let\PYGdefault@bf=\textbf\def\PYGdefault@tc##1{\textcolor[rgb]{0.00,0.50,0.00}{##1}}}
\expandafter\def\csname PYGdefault@tok@na\endcsname{\def\PYGdefault@tc##1{\textcolor[rgb]{0.49,0.56,0.16}{##1}}}
\expandafter\def\csname PYGdefault@tok@c1\endcsname{\let\PYGdefault@it=\textit\def\PYGdefault@tc##1{\textcolor[rgb]{0.25,0.50,0.50}{##1}}}
\expandafter\def\csname PYGdefault@tok@ne\endcsname{\let\PYGdefault@bf=\textbf\def\PYGdefault@tc##1{\textcolor[rgb]{0.82,0.25,0.23}{##1}}}
\expandafter\def\csname PYGdefault@tok@nn\endcsname{\let\PYGdefault@bf=\textbf\def\PYGdefault@tc##1{\textcolor[rgb]{0.00,0.00,1.00}{##1}}}
\expandafter\def\csname PYGdefault@tok@mf\endcsname{\def\PYGdefault@tc##1{\textcolor[rgb]{0.40,0.40,0.40}{##1}}}
\expandafter\def\csname PYGdefault@tok@kd\endcsname{\let\PYGdefault@bf=\textbf\def\PYGdefault@tc##1{\textcolor[rgb]{0.00,0.50,0.00}{##1}}}
\expandafter\def\csname PYGdefault@tok@s2\endcsname{\def\PYGdefault@tc##1{\textcolor[rgb]{0.73,0.13,0.13}{##1}}}
\expandafter\def\csname PYGdefault@tok@cp\endcsname{\def\PYGdefault@tc##1{\textcolor[rgb]{0.74,0.48,0.00}{##1}}}
\expandafter\def\csname PYGdefault@tok@gh\endcsname{\let\PYGdefault@bf=\textbf\def\PYGdefault@tc##1{\textcolor[rgb]{0.00,0.00,0.50}{##1}}}
\expandafter\def\csname PYGdefault@tok@mo\endcsname{\def\PYGdefault@tc##1{\textcolor[rgb]{0.40,0.40,0.40}{##1}}}
\expandafter\def\csname PYGdefault@tok@fm\endcsname{\def\PYGdefault@tc##1{\textcolor[rgb]{0.00,0.00,1.00}{##1}}}
\expandafter\def\csname PYGdefault@tok@s\endcsname{\def\PYGdefault@tc##1{\textcolor[rgb]{0.73,0.13,0.13}{##1}}}
\expandafter\def\csname PYGdefault@tok@sx\endcsname{\def\PYGdefault@tc##1{\textcolor[rgb]{0.00,0.50,0.00}{##1}}}
\expandafter\def\csname PYGdefault@tok@vc\endcsname{\def\PYGdefault@tc##1{\textcolor[rgb]{0.10,0.09,0.49}{##1}}}
\expandafter\def\csname PYGdefault@tok@w\endcsname{\def\PYGdefault@tc##1{\textcolor[rgb]{0.73,0.73,0.73}{##1}}}
\expandafter\def\csname PYGdefault@tok@nd\endcsname{\def\PYGdefault@tc##1{\textcolor[rgb]{0.67,0.13,1.00}{##1}}}
\expandafter\def\csname PYGdefault@tok@mh\endcsname{\def\PYGdefault@tc##1{\textcolor[rgb]{0.40,0.40,0.40}{##1}}}
\expandafter\def\csname PYGdefault@tok@gr\endcsname{\def\PYGdefault@tc##1{\textcolor[rgb]{1.00,0.00,0.00}{##1}}}

\makeatother

\makeatletter
\def\PYG@reset{\let\PYG@it=\relax \let\PYG@bf=\relax%
    \let\PYG@ul=\relax \let\PYG@tc=\relax%
    \let\PYG@bc=\relax \let\PYG@ff=\relax}
\def\PYG@tok#1{\csname PYG@tok@#1\endcsname}
\def\PYG@toks#1+{\ifx\relax#1\empty\else%
    \PYG@tok{#1}\expandafter\PYG@toks\fi}
\def\PYG@do#1{\PYG@bc{\PYG@tc{\PYG@ul{%
    \PYG@it{\PYG@bf{\PYG@ff{#1}}}}}}}
\def\PYG#1#2{\PYG@reset\PYG@toks#1+\relax+\PYG@do{#2}}

\expandafter\def\csname PYG@tok@kt\endcsname{\def\PYG@tc##1{\textcolor[rgb]{0.69,0.00,0.25}{##1}}}
\expandafter\def\csname PYG@tok@dl\endcsname{\def\PYG@tc##1{\textcolor[rgb]{0.73,0.13,0.13}{##1}}}
\expandafter\def\csname PYG@tok@vg\endcsname{\def\PYG@tc##1{\textcolor[rgb]{0.10,0.09,0.49}{##1}}}
\expandafter\def\csname PYG@tok@nf\endcsname{\def\PYG@tc##1{\textcolor[rgb]{0.00,0.00,1.00}{##1}}}
\expandafter\def\csname PYG@tok@nc\endcsname{\let\PYG@bf=\textbf\def\PYG@tc##1{\textcolor[rgb]{0.00,0.00,1.00}{##1}}}
\expandafter\def\csname PYG@tok@nv\endcsname{\def\PYG@tc##1{\textcolor[rgb]{0.10,0.09,0.49}{##1}}}
\expandafter\def\csname PYG@tok@cp\endcsname{\def\PYG@tc##1{\textcolor[rgb]{0.74,0.48,0.00}{##1}}}
\expandafter\def\csname PYG@tok@nl\endcsname{\def\PYG@tc##1{\textcolor[rgb]{0.63,0.63,0.00}{##1}}}
\expandafter\def\csname PYG@tok@sc\endcsname{\def\PYG@tc##1{\textcolor[rgb]{0.73,0.13,0.13}{##1}}}
\expandafter\def\csname PYG@tok@w\endcsname{\def\PYG@tc##1{\textcolor[rgb]{0.73,0.73,0.73}{##1}}}
\expandafter\def\csname PYG@tok@vm\endcsname{\def\PYG@tc##1{\textcolor[rgb]{0.10,0.09,0.49}{##1}}}
\expandafter\def\csname PYG@tok@cpf\endcsname{\let\PYG@it=\textit\def\PYG@tc##1{\textcolor[rgb]{0.25,0.50,0.50}{##1}}}
\expandafter\def\csname PYG@tok@ne\endcsname{\let\PYG@bf=\textbf\def\PYG@tc##1{\textcolor[rgb]{0.82,0.25,0.23}{##1}}}
\expandafter\def\csname PYG@tok@cs\endcsname{\let\PYG@it=\textit\def\PYG@tc##1{\textcolor[rgb]{0.25,0.50,0.50}{##1}}}
\expandafter\def\csname PYG@tok@go\endcsname{\def\PYG@tc##1{\textcolor[rgb]{0.53,0.53,0.53}{##1}}}
\expandafter\def\csname PYG@tok@ni\endcsname{\let\PYG@bf=\textbf\def\PYG@tc##1{\textcolor[rgb]{0.60,0.60,0.60}{##1}}}
\expandafter\def\csname PYG@tok@ss\endcsname{\def\PYG@tc##1{\textcolor[rgb]{0.10,0.09,0.49}{##1}}}
\expandafter\def\csname PYG@tok@sx\endcsname{\def\PYG@tc##1{\textcolor[rgb]{0.00,0.50,0.00}{##1}}}
\expandafter\def\csname PYG@tok@ch\endcsname{\let\PYG@it=\textit\def\PYG@tc##1{\textcolor[rgb]{0.25,0.50,0.50}{##1}}}
\expandafter\def\csname PYG@tok@no\endcsname{\def\PYG@tc##1{\textcolor[rgb]{0.53,0.00,0.00}{##1}}}
\expandafter\def\csname PYG@tok@gu\endcsname{\let\PYG@bf=\textbf\def\PYG@tc##1{\textcolor[rgb]{0.50,0.00,0.50}{##1}}}
\expandafter\def\csname PYG@tok@sr\endcsname{\def\PYG@tc##1{\textcolor[rgb]{0.73,0.40,0.53}{##1}}}
\expandafter\def\csname PYG@tok@gp\endcsname{\let\PYG@bf=\textbf\def\PYG@tc##1{\textcolor[rgb]{0.00,0.00,0.50}{##1}}}
\expandafter\def\csname PYG@tok@nn\endcsname{\let\PYG@bf=\textbf\def\PYG@tc##1{\textcolor[rgb]{0.00,0.00,1.00}{##1}}}
\expandafter\def\csname PYG@tok@sb\endcsname{\def\PYG@tc##1{\textcolor[rgb]{0.73,0.13,0.13}{##1}}}
\expandafter\def\csname PYG@tok@kd\endcsname{\let\PYG@bf=\textbf\def\PYG@tc##1{\textcolor[rgb]{0.00,0.50,0.00}{##1}}}
\expandafter\def\csname PYG@tok@mi\endcsname{\def\PYG@tc##1{\textcolor[rgb]{0.40,0.40,0.40}{##1}}}
\expandafter\def\csname PYG@tok@m\endcsname{\def\PYG@tc##1{\textcolor[rgb]{0.40,0.40,0.40}{##1}}}
\expandafter\def\csname PYG@tok@s2\endcsname{\def\PYG@tc##1{\textcolor[rgb]{0.73,0.13,0.13}{##1}}}
\expandafter\def\csname PYG@tok@mh\endcsname{\def\PYG@tc##1{\textcolor[rgb]{0.40,0.40,0.40}{##1}}}
\expandafter\def\csname PYG@tok@gs\endcsname{\let\PYG@bf=\textbf}
\expandafter\def\csname PYG@tok@mb\endcsname{\def\PYG@tc##1{\textcolor[rgb]{0.40,0.40,0.40}{##1}}}
\expandafter\def\csname PYG@tok@si\endcsname{\let\PYG@bf=\textbf\def\PYG@tc##1{\textcolor[rgb]{0.73,0.40,0.53}{##1}}}
\expandafter\def\csname PYG@tok@na\endcsname{\def\PYG@tc##1{\textcolor[rgb]{0.49,0.56,0.16}{##1}}}
\expandafter\def\csname PYG@tok@kc\endcsname{\let\PYG@bf=\textbf\def\PYG@tc##1{\textcolor[rgb]{0.00,0.50,0.00}{##1}}}
\expandafter\def\csname PYG@tok@ow\endcsname{\let\PYG@bf=\textbf\def\PYG@tc##1{\textcolor[rgb]{0.67,0.13,1.00}{##1}}}
\expandafter\def\csname PYG@tok@se\endcsname{\let\PYG@bf=\textbf\def\PYG@tc##1{\textcolor[rgb]{0.73,0.40,0.13}{##1}}}
\expandafter\def\csname PYG@tok@gd\endcsname{\def\PYG@tc##1{\textcolor[rgb]{0.63,0.00,0.00}{##1}}}
\expandafter\def\csname PYG@tok@k\endcsname{\let\PYG@bf=\textbf\def\PYG@tc##1{\textcolor[rgb]{0.00,0.50,0.00}{##1}}}
\expandafter\def\csname PYG@tok@gr\endcsname{\def\PYG@tc##1{\textcolor[rgb]{1.00,0.00,0.00}{##1}}}
\expandafter\def\csname PYG@tok@c\endcsname{\let\PYG@it=\textit\def\PYG@tc##1{\textcolor[rgb]{0.25,0.50,0.50}{##1}}}
\expandafter\def\csname PYG@tok@sd\endcsname{\let\PYG@it=\textit\def\PYG@tc##1{\textcolor[rgb]{0.73,0.13,0.13}{##1}}}
\expandafter\def\csname PYG@tok@s\endcsname{\def\PYG@tc##1{\textcolor[rgb]{0.73,0.13,0.13}{##1}}}
\expandafter\def\csname PYG@tok@vc\endcsname{\def\PYG@tc##1{\textcolor[rgb]{0.10,0.09,0.49}{##1}}}
\expandafter\def\csname PYG@tok@mo\endcsname{\def\PYG@tc##1{\textcolor[rgb]{0.40,0.40,0.40}{##1}}}
\expandafter\def\csname PYG@tok@fm\endcsname{\def\PYG@tc##1{\textcolor[rgb]{0.00,0.00,1.00}{##1}}}
\expandafter\def\csname PYG@tok@cm\endcsname{\let\PYG@it=\textit\def\PYG@tc##1{\textcolor[rgb]{0.25,0.50,0.50}{##1}}}
\expandafter\def\csname PYG@tok@nt\endcsname{\let\PYG@bf=\textbf\def\PYG@tc##1{\textcolor[rgb]{0.00,0.50,0.00}{##1}}}
\expandafter\def\csname PYG@tok@nb\endcsname{\def\PYG@tc##1{\textcolor[rgb]{0.00,0.50,0.00}{##1}}}
\expandafter\def\csname PYG@tok@sh\endcsname{\def\PYG@tc##1{\textcolor[rgb]{0.73,0.13,0.13}{##1}}}
\expandafter\def\csname PYG@tok@kn\endcsname{\let\PYG@bf=\textbf\def\PYG@tc##1{\textcolor[rgb]{0.00,0.50,0.00}{##1}}}
\expandafter\def\csname PYG@tok@gt\endcsname{\def\PYG@tc##1{\textcolor[rgb]{0.00,0.27,0.87}{##1}}}
\expandafter\def\csname PYG@tok@ge\endcsname{\let\PYG@it=\textit}
\expandafter\def\csname PYG@tok@o\endcsname{\def\PYG@tc##1{\textcolor[rgb]{0.40,0.40,0.40}{##1}}}
\expandafter\def\csname PYG@tok@sa\endcsname{\def\PYG@tc##1{\textcolor[rgb]{0.73,0.13,0.13}{##1}}}
\expandafter\def\csname PYG@tok@bp\endcsname{\def\PYG@tc##1{\textcolor[rgb]{0.00,0.50,0.00}{##1}}}
\expandafter\def\csname PYG@tok@mf\endcsname{\def\PYG@tc##1{\textcolor[rgb]{0.40,0.40,0.40}{##1}}}
\expandafter\def\csname PYG@tok@gi\endcsname{\def\PYG@tc##1{\textcolor[rgb]{0.00,0.63,0.00}{##1}}}
\expandafter\def\csname PYG@tok@gh\endcsname{\let\PYG@bf=\textbf\def\PYG@tc##1{\textcolor[rgb]{0.00,0.00,0.50}{##1}}}
\expandafter\def\csname PYG@tok@vi\endcsname{\def\PYG@tc##1{\textcolor[rgb]{0.10,0.09,0.49}{##1}}}
\expandafter\def\csname PYG@tok@il\endcsname{\def\PYG@tc##1{\textcolor[rgb]{0.40,0.40,0.40}{##1}}}
\expandafter\def\csname PYG@tok@kr\endcsname{\let\PYG@bf=\textbf\def\PYG@tc##1{\textcolor[rgb]{0.00,0.50,0.00}{##1}}}
\expandafter\def\csname PYG@tok@c1\endcsname{\let\PYG@it=\textit\def\PYG@tc##1{\textcolor[rgb]{0.25,0.50,0.50}{##1}}}
\expandafter\def\csname PYG@tok@kp\endcsname{\def\PYG@tc##1{\textcolor[rgb]{0.00,0.50,0.00}{##1}}}
\expandafter\def\csname PYG@tok@err\endcsname{\def\PYG@bc##1{\setlength{\fboxsep}{0pt}\fcolorbox[rgb]{1.00,0.00,0.00}{1,1,1}{\strut ##1}}}
\expandafter\def\csname PYG@tok@nd\endcsname{\def\PYG@tc##1{\textcolor[rgb]{0.67,0.13,1.00}{##1}}}
\expandafter\def\csname PYG@tok@s1\endcsname{\def\PYG@tc##1{\textcolor[rgb]{0.73,0.13,0.13}{##1}}}

\def\PYGZus{\char`\_}

\def\PYGZsh{\char`\#}

\makeatother

\usepackage{enumitem}

\graphicspath{ {./fig/} }

\title{A Survey and Implementation of Performance Metrics for Self-Organized Maps}

\author{%
  Florent~Forest \\
  LIPN, Université Sorbonne Paris Nord \\
  Safran Aircraft Engines \\
  \texttt{forest@lipn.univ-paris13.fr}
  \And
  Mustapha~Lebbah \\
  LIPN, Université Sorbonne Paris Nord \\
  \texttt{lebbah@lipn.univ-paris13.fr}
  \And
  Hanane~Azzag \\
  LIPN, Université Sorbonne Paris Nord \\
  \texttt{azzag@lipn.univ-paris13.fr}
  \And
  Jérôme~Lacaille \\
  Safran Aircraft Engines \\
  \texttt{jerome.lacaille@safrangroup.com}
}

\begin{document}

\maketitle

\begin{abstract} 
  Self-Organizing Map algorithms have been used for almost 40 years across various application domains such as biology, geology, healthcare, industry and humanities as an interpretable tool to explore, cluster and visualize high-dimensional data sets. In every application, practitioners need to know whether they can \textit{trust} the resulting mapping, and perform model selection to tune algorithm parameters (e.g. the map size). Quantitative evaluation of self-organizing maps (SOM) is a subset of clustering validation, which is a challenging problem as such. Clustering model selection is typically achieved by using clustering validity indices. While they also apply to self-organized clustering models, they ignore the topology of the map, only answering the question: do the SOM code vectors approximate well the data distribution? Evaluating SOM models brings in the additional challenge of assessing their topology: does the mapping preserve neighborhood relationships between the map and the original data? The problem of assessing the performance of SOM models has already been tackled quite thoroughly in literature, giving birth to a family of quality indices incorporating neighborhood constraints, called \textit{topographic} indices. Commonly used examples of such metrics are the topographic error, neighborhood preservation or the topographic product. However, open-source implementations are almost impossible to find. This is the issue we try to solve in this work: after a survey of existing SOM performance metrics, we implemented them in Python and widely used numerical libraries, and provide them as an open-source library, SOMperf\footnote{\url{https://github.com/FlorentF9/SOMperf}}. This paper introduces each metric available in our module along with usage examples.
\end{abstract}

\section{Introduction}

Self-Organizing Maps (SOM) \cite{Kohonen1982} have been used for almost 40 years across various application domains such as biology, geology, healthcare, industry \cite{Come2011,Forest2020} and humanities \cite{Massoni2009} as an interpretable tool to explore, cluster and visualize high-dimensional data sets. In every application, practitioners need to know whether they can \textit{trust} the resulting mapping, and perform model selection to select algorithm parameters (e.g. the map size, learning rate and number of iterations). Concretely, two questions need to be answered:
\begin{itemize}[noitemsep]
    \item Do the SOM code vectors approximate well the data distribution?
    \item Does the mapping preserve neighborhood relationships between the map and the original data space?
\end{itemize}
Quantitative evaluation of SOM is a subset of clustering validation, which is a challenging problem, for the obvious reason that there generally is no ground truth against which results could be tested, unlike in supervised classification. Clustering model selection is typically achieved using validity indices. While these also apply to SOM, they ignore the topology of the map, only answering the first question. The second question brings in the additional challenge of assessing their topology. The problem of assessing SOM performance has already been tackled quite thoroughly in literature, giving birth to a family of quality indices incorporating neighborhood constraints, qualified as \textit{topographic} indices \cite{Kiviluoto1996,Polzlbauer2004}. However, while open-source software for SOM are available in various languages (Matlab \cite{somtoolbox1999}, R \cite{sombrero2014}, Python \cite{sompy2014}, Scala with distributed versions using Spark \cite{sparkmlsom2019}, etc.), implementations for quality indices are almost impossible to find. This is the issue we aim to solve in this work: after a survey of existing SOM performance metrics, we implemented them in Python, one of the most popular languages for data mining today, and provide them as an open-source library, SOMperf, available at \url{https://github.com/FlorentF9/SOMperf}. 

The rest of this paper is structured as follows: first, a brief background on SOM provides the necessary concepts and notations. Then, each metric available in our library is introduced. Finally, we provide some usage examples.

\section{Background on SOM}

The Self-Organizing Map (SOM) \cite{Kohonen1982} is a clustering model that introduces a topological relationship between clusters. It consists in a network of two layers: an input layer, and an output layer of interconnected nodes, often called \emph{neurons} or \emph{units}. Typically, the topology of this layer is chosen as a two-dimensional grid, because it can be easily visualized. This visualization capability characterizes SOM as an interpretable clustering method. In other words, a SOM approximates a data distribution by a lower-dimensional \emph{manifold}. The set of input data samples is denoted $\mathbb{X} = \{\mathbf{x}_i\}_{1 \leq i \leq N}, \mathbf{x}_i \in \mathbb{R}^D$. A SOM is composed of $K$ units, associated to the set of prototype vectors $\{\mathbf{m}_k\}_{1 \leq k \leq K} \in \mathbb{R}^D$. A data point is projected on the map by finding its closest prototype according to euclidean distance. The corresponding map unit is called \textit{best-matching unit} (BMU). We introduce the notation $b_i$ for the BMU of $\mathbf{x}_i$: 
\begin{equation}
    b_i = \underset{k}{\text{argmin}}||\mathbf{x}_i-\mathbf{m}_k||_2^2.
\end{equation}
The grid topology allows to define an inter-node distance $\delta(k,l)$,  which is generally the length of the shortest path between units $k$ and $l$ on the map. We then define the neighborhood function and a temperature parameter $T$, controlling the radius of the neighborhood around a unit. A common choice is a Gaussian neighborhood function $\mathcal{K}^T(d) = e^{-d^2 / T^2}$, where $T$ is decreased at each training iteration. The original SOM learning algorithm, also called \textit{stochastic algorithm} or \textit{Kohonen algorithm}, takes each training sample $\mathbf{x}_i$ and updates every prototype vector by moving them closer to the point $\mathbf{x}_i$. The updates are weighted by the neighborhood around the BMU, so that neighboring units receive a large update and very distant units are not updated at all. This expresses as the following update rule:
\begin{equation}
    \mathbf{m}_k \gets \mathbf{m}_k + \alpha \mathcal{K}^T\left(\delta(b_i, k)\right) (\mathbf{x}_i - \mathbf{m}_k) \label{eq:som}
\end{equation}
where $\alpha$ is a learning rate that is decreased during training.
The stochastic algorithm is detailed in Algorithm~\ref{alg:stochastic-som}. A disadvantage of this algorithm is that it converges slowly, is sequential and cannot be parallelized. Therefore, another algorithm was introduced: the batch SOM algorithm. It consists in minimizing a cost function called \textit{distortion} (see Equation~\ref{eq:distortion} in the next section) in a $K$-means-like alternate fashion.
\begin{algorithm}
    \KwIn{training set $\mathbb{X}$; SOM map size; temperatures $T_{max}$, $T_{min}$; $\mathit{iterations}$}
    \KwOut{SOM code vectors $\{\mathbf{m}_k\}$}
    Initialize SOM parameters $\{\mathbf{m}_k\}$ \;
    \For{$\mathit{n} = 1, \dots, \mathit{iterations}$}{
        $T \gets T_{max} \left(\frac{T_{min}}{T_{max}}\right)^{\mathit{n}/\mathit{iterations}}$ \;
        Load next training sample $\mathbf{x}_i$ \;
        Compute BMU $b_i$ \;
        \For{$\mathit{k} = 1, \dots, K$}{
            Update prototype $\mathbf{m}_k$ (by equation~\ref{eq:som}) \;
        }
    }
    \caption{Stochastic SOM algorithm.}\label{alg:stochastic-som}
\end{algorithm}

\section{Measuring SOM performance}

\begin{figure}
    \centering
    \includegraphics[width=0.9\textwidth]{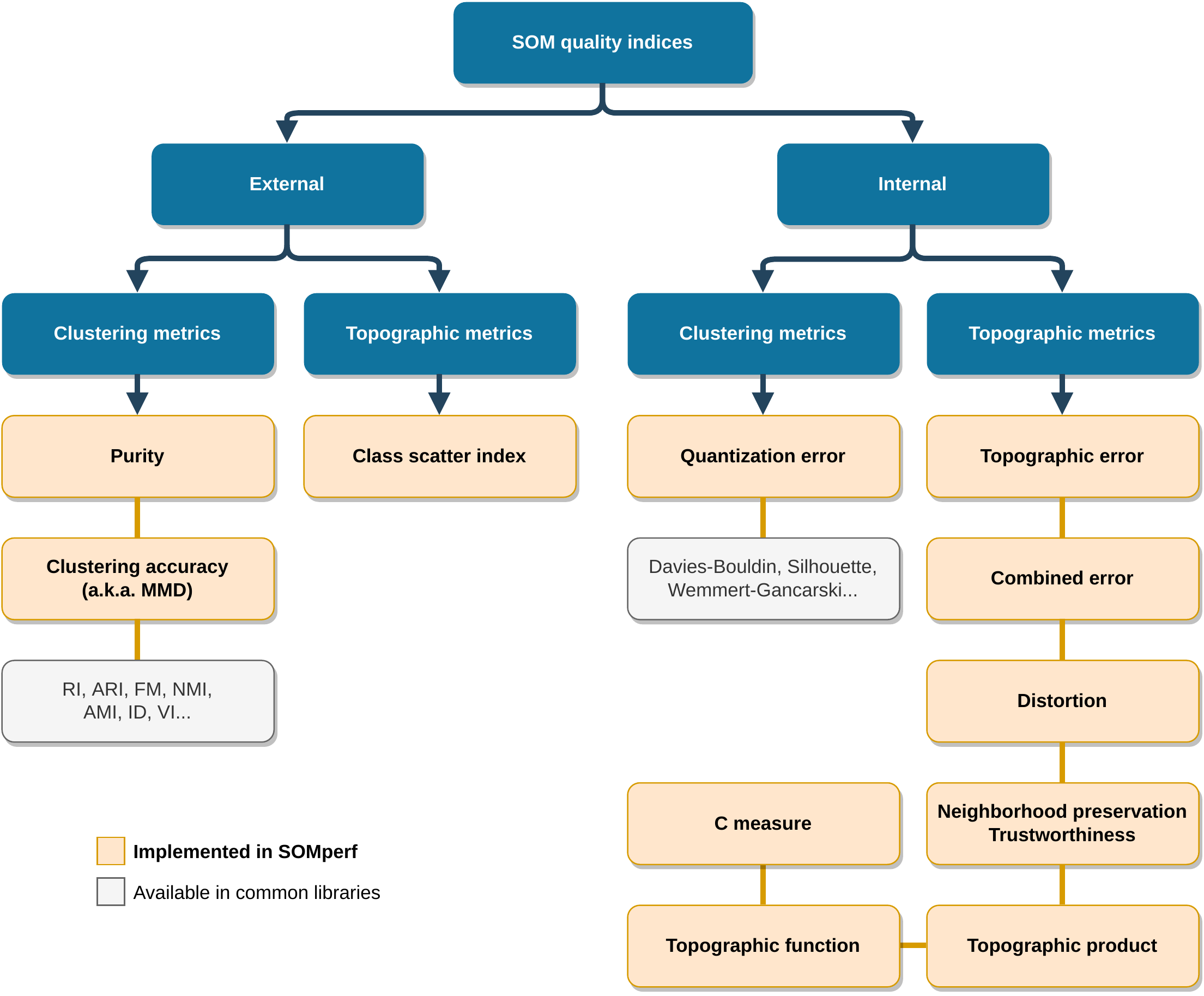}
    \caption{SOM performance metrics can be classified into external (label-based) or internal indices, and based on whether they evaluate topology (topographic metrics) or not (clustering metrics).}
    \label{fig:som-indices}
\end{figure}

An overview of all metrics implemented in SOMperf is depicted in Figure~\ref{fig:som-indices}. These can be categorized into two families:
\begin{enumerate}[noitemsep]
    \item Clustering metrics. Any clustering quality measure that relies solely on the prototype vectors and not on their topological organization. This encompasses all quality indices used in clustering literature.
    \item Topographic metrics. Under this term, we coin quality measures that, on the contrary, assess the topological organization of the model. Some indices also evaluate the clustering quality, but we call it topographic as soon as it incorporates the map topology. In particular, they must detect neighborhood violations such as \emph{foldings}.
\end{enumerate}
On another level, we can classify them into two well-known families, depending on the use of ground-truth labels:
\begin{enumerate}[noitemsep]
    \item Internal indices, using only intrinsic properties of the model and the data. 
    \item External indices, relying on external ground-truth class labels to evaluate results.
\end{enumerate}
For instance, quantization error falls into the clustering metric category (as it measures how SOM cluster centers fit the data distribution, without using any topology information) and is an internal index (not depending on external labels). On the other side, the Class Scatter Index is a topographic metric and an external index, as it measures how ground-truth class labels are organized into groups of neighboring map units.
Internal clustering indices measure the quality of a clustering is absence of ground-truth labels, which is mostly the case in unsupervised data exploration.
The majority of internal criteria rely on a combination of between-cluster and within-cluster distances: between-cluster distance measures how distinct clusters are dissimilar, while within-cluster distance measures how elements belonging to the same cluster are similar.
See for instance \cite{Arbelaitz2013} or \cite{Hamalainen2017} for recent reviews. External indices compute a similarity (or distance) between two partitions. They are used either to compare a clustering with ground-truth class labels, or to compare two clusterings.
External indices are of two kinds: first, count-based indices, based on counting how many pairs of points belong to the same or to different clusters in each partition (using a contingency matrix). Second, information theoretic measures, measuring how much information is provided by knowing one partition on the second partition.
Commonly used indices are the adjusted Rand index (ARI) and information theoretic measures such as Mutual Information and its normalized (NMI), adjusted (AMI) and standardized (SMI) variants. Numerous studies have been devoted to external indices, see for instance \cite{Vinh2010}.

\subsection{Internal indices}

\paragraph{Quantization error} 

Quantization error is the average error made by projecting data on the SOM, as measured by euclidean distance, i.e. the mean euclidean distance between a data sample and its best-matching unit:
\begin{equation}
    \text{QE}(\{\mathbf{m}_k\}, \mathbb{X}) = \frac{1}{N} \sum_{i=1}^N ||\mathbf{x}_i - \mathbf{m}_{b_i}||_2.
\end{equation}

\paragraph{Distortion}

Distortion is the loss minimized by the SOM learning algorithm. It is similar to the within-cluster sum of squared errors minimized by $K$-means, but with a topology constraint. It is calculated by the sum of squared euclidean distances between each map prototype and sample, weighted by the neighborhood function:
\begin{equation}
    \text{D}(\{\mathbf{m}_k\}, \mathbb{X}, T) = \frac{1}{N} \sum_{i=1}^N \sum_{k=1}^K  \mathcal{K}^T\left(\delta(b_i, k)\right) ||\mathbf{x}_i - \mathbf{m}_k||_2^2.
    \label{eq:distortion}
\end{equation}

\paragraph{Topographic error}

Topographic error \cite{Kiviluoto1996} assesses the self-organization of a SOM model. It is calculated as the fraction of samples whose best and second-best matching units are not neighbors on the map. In other words, this error quantifies the smoothness of projections on the self-organized map. Using the notation $b^k_i$ for the $k$-th best-matching units of $\mathbf{x}_i$, we define the topographic error:
\begin{equation}
    \text{TE}(\{\mathbf{m}_k\}, \mathbb{X}) = \frac{1}{N} \sum_{i=1}^N \mathbbm{1}_{\delta(b^1_i, b^2_i) > 1}.
\end{equation}

\paragraph{Combined error}

Combined error \cite{Kaski1996} is an error measure that combines and extends quantification and topographic errors. Its computation is more complex than the previous indices. For a given data sample $\mathbf{x}_i$, we first compute its two best matching units $b^1_i$ and $b^2_i$. Then, we compute a sum of euclidean distances from $\mathbf{x}_i$ to the second BMU's prototype vector $\mathbf{m}_{b^2_i}$, starting with the distance from $\mathbf{x}_i$ to $\mathbf{m}_{b^1_i}$, and thereafter following a shortest path until $\mathbf{m}_{b^2_i}$, \textit{going only through neighboring units on the map}. Let $p$ be a path on the map of length $P \geq 1$, from $p(0) = b^1_i$ to $p(L) = b^2_i$, such that $p(k)$ and $p(k+1)$ must be neighbors for $k = 0 \ldots P-1$. The distance along the shortest path on the map is computed as
\begin{equation}
    \text{CE}_i = ||\mathbf{x}_i -  \mathbf{m}_{b^1_i}||_2^2 + \underset{p}{\min} \sum_{k=0}^{P-1} ||\mathbf{m}_{p(k+1)} - \mathbf{m}_{p(k)}||_2^2.
\end{equation}
Finally, combined error (CE) is the average of this distance over the input samples:
\begin{equation}
    \text{CE}(\{\mathbf{m}_k\}, \mathbb{X}) = \frac{1}{N} \sum_{i=1}^N \text{CE}_i.
\end{equation}

\paragraph{Neighborhood preservation and trustworthiness}

\begin{figure}
    \centering
    \includegraphics[width=0.6\textwidth]{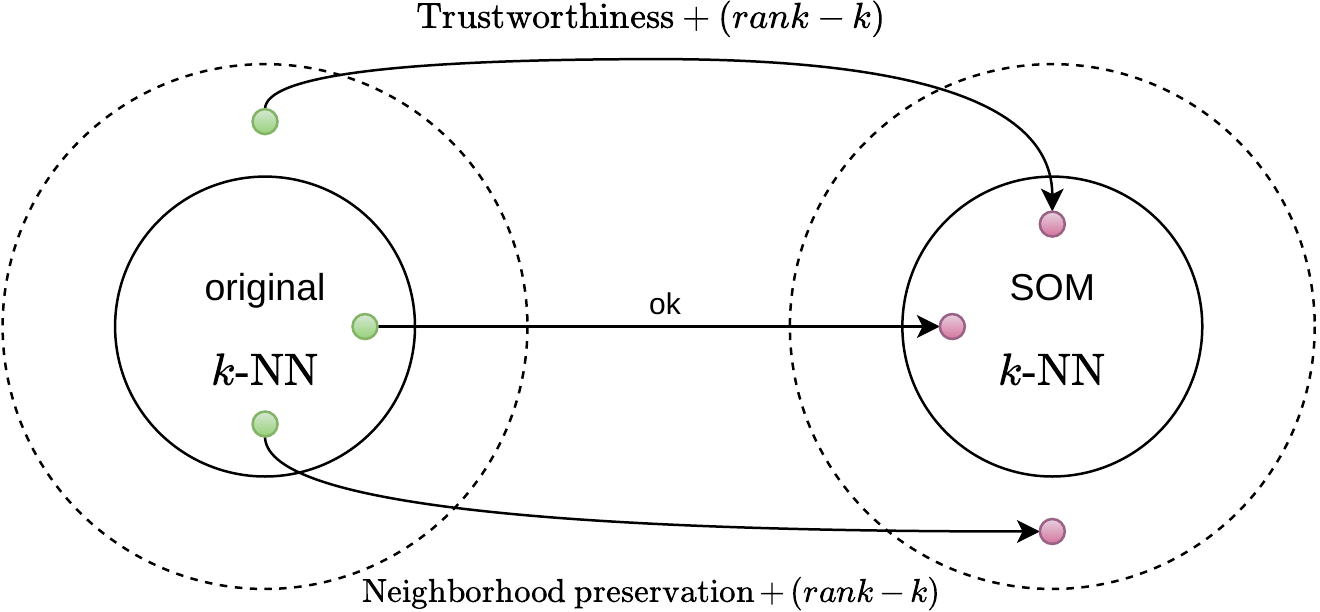}
    \caption{Illustration of neighborhood preservation and trustworthiness measures.}\label{fig:np-trustworthiness}
\end{figure}

The neighborhood preservation and trustworthiness \cite{Venna2001} measure how the projection preserves neighborhoods in the input (respectively output) space by ranking the $k$-nearest neighbors ($k$-NN) of each sample before and after projection. For a given $k$, each sample contributes negatively by the difference between its rank and $k$. See Figure~\ref{fig:np-trustworthiness} for an illustration.
An important issue with implementing neighborhood preservation and trustworthiness is handling ties in the projected $k$-NN. As the SOM projections are discrete, all samples projected onto the same map unit will have a distance equal to 0. In the continuous input space, exact ties are very unlikely but still possible. Four methods to tackle this problem are exposed in \cite{Polzlbauer2004}. We adopt a weighted averaging approach as it is reasonable and deterministic. We include all ties in the set of $k$-NN, possibly producing more than $k$ neighbors (in particular in the discrete projected space). To stay in the $[0, 1]$ range, every error term is weighted by:
\begin{itemize}[noitemsep]
 \item For trustworthiness: the ratio between the number of elements in the original $k$-NN (most often, exactly $k$), and the number of elements in the projected $k$-NN (most often, larger than $k$).
 \item For neighborhood preservation: the inverse of this ratio.
\end{itemize}

\paragraph{Topographic product}

The topographic product (TP) \cite{Bauer1992a} measures the preservation of neighborhood relations between input space and the map. It depends only on the prototype vectors and map topology, and is able to indicate whether the dimension of the map is appropriate to fit the data set, or if it introduced neighborhood violations, induced by \emph{foldings} of the map. We note $d$ the euclidean distance in input space, and $\delta$ the topographic distance on the map. The computation of TP starts by defining two ratios between the distance of a prototype $j$ to its $k$-th nearest neighbor on the map $n_k^{\delta}(j)$, and to its $k$-th nearest neighbor in input space $n_k^d(j)$:
\begin{equation*}
    Q_1(j, k) = \frac{d\left( \mathbf{m}_j, \mathbf{m}_{n_k^{\delta}(j)} \right)}{d\left( \mathbf{m}_j, \mathbf{m}_{n_k^d(j)} \right)}, \; Q_2(j, k) = \frac{\delta\left(j, n_k^{\delta}(j)\right)}{\delta\left(j, n_k^d(j)\right)}.
\end{equation*}
Naturally, we always have $Q_1 \geq 1$ and $Q_2 \leq 1$. The ratios are combined into a product in order to obtain a symmetric measure and mitigate local magnification factors:
\begin{equation*}
    P_3(j, k) = \left[ \prod_{l=1}^k Q_1(j, l) Q_2(j, l) \right]^{\frac{1}{2k}}.
\end{equation*}
Finally, TP is obtained by taking the logarithm and averaging over all map units and neighborhood orders:
\begin{equation}
    \text{TP}(\{\mathbf{m}_k\}) = \frac{1}{K(K-1)}\sum_{j=1}^K \sum_{k=1}^{K-1} \log P_3(j, k).
\end{equation}
$\text{TP} < 0$ indicates the map dimension is too low to correctly represent the data set; $\text{TP} = 0$ means the dimension is adequate; and $\text{TP} > 0$ indicates a dimension too high and neighborhood violations. However, as the TP only uses the map prototypes, it is unable to distinguish between foldings of a non-linear data manifold, and foldings due to neighborhood violations. As a consequence, it is limited to linear data manifolds.

\paragraph{Topographic function}

The topographic function (TF) \cite{Villmann1994} intends to overcome the limitation of the topographic product, by distinguishing between natural foldings of a non-linear data manifold and incorrect foldings due to neighborhood violations. The main difference is that is uses not only the prototype vectors, but the receptive fields of each unit $c$, defined as $R_c = \{ \mathbf{x} \in \mathbb{X} \; | \; \underset{k}{\text{argmin}}||\mathbf{x}-\mathbf{m}_k||_2^2 = c\}$. For each unit $c$ and integer $k \in \{1, \ldots, K\}$, the number of units having adjacent receptive fields and a distance to $c$ on the map larger than $k$, is computed:
\begin{equation*}
    f_c(k) = | \{c' \in \{1, \ldots, K\} \; | \; R_c \cap R_{c'} \neq \emptyset \; \land \; \delta(c, c') > k \} |.
\end{equation*}
The topographic function is defined by summing over all map units:
\begin{equation}
    \text{TF}(k) = \sum_{c=1}^K f_c(k).
\end{equation}
A normalization is necessary to compare maps of different sizes, replacing $k$ by $k/\delta_{\text{max}}$ (the maximum distance on the map) and dividing by $K(K-3^p)$ where $p$ is the number of dimensions of the SOM (generally $p=2$). In practice, the receptive fields can be easily estimated without computing the full Voronoi tesselation, by building a connectivity matrix connecting each pair of units that are the BMU and second-BMU of a given data point.

\paragraph{Kruskal-Shepard error}

Kruskal-Shepard error, introduced for multi-dimensional scaling \cite{Kruskal1964}, measures the preservation of pairwise distances between two different spaces, and was used for the SOM in \cite{Elend2019}. It is computed by the squared Frobenius norm between the pairwise distance matrix of the dataset, and the distance matrix betwen units on the SOM. Both distance matrices are scaled to the $[0, 1]$ range. We consider the normalized error, by dividing it by $N(N -1)$:
\begin{align*}
    \text{KSE}(\{\mathbf{m}_k\}, \mathbb{X})& = \frac{1}{N(N -1)} ||\mathbf{D}^{\mathbb{X}} - \mathbf{D}^{\text{SOM}}||_F^2 \\
    \text{where } \mathbf{D}^{\mathbb{X}}_{ij} &= \frac{||\mathbf{x}_i - \mathbf{x}_j||_2^2}{\underset{i',j'}{\max}\;||\mathbf{x}_{i'} - \mathbf{x}_{j'}||_2^2} \\
    \text{and } \mathbf{D}^{\text{SOM}}_{ij} &= \frac{\delta(b_i, b_j)}{\underset{k,l}{\max}\;\delta(k, l)}.
\end{align*}

\paragraph{C measure}

The C measure \cite{Goodhill1996} measures neighborhood preservation between two spaces similarly to the Kruskal-Shepard error, but using the element-wise products between pairwise distances in the input and output space:
\begin{equation}
    \text{C}(\{\mathbf{m}_k\}, \mathbb{X}) = \sum_{i=1}^N \sum_{j<i} d(\mathbf{x}_i - \mathbf{x}_j) \; \delta(b_i, b_j).
\end{equation}
This quantity must be maximized, meaning that both distances must take large values at the same time, or in other words, if two points are far apart in the input space, they must also be far apart in the output space.

\subsection{External indices}

A clustering into $K$ clusters is described by the sets of data points belonging to each cluster, noted $\mathbf{Q} = \{Q_k\}, k = 1 \ldots K$. In order to define external indices, we assume labels are associated to each sample, with to $C$ different classes. We note $\mathbf{Y} = \{Y_j\}, j = 1 \ldots C$ the sets of elements belonging to each class.

\paragraph{Purity}

Purity measures the purity of clusters with respect to ground-truth class labels. To compute the purity of a clustering $\mathbf{Q}$, each cluster is assigned to the class which is most frequent in the cluster, and then the accuracy of this assignment is measured by counting the number of correctly assigned points and dividing by the total number of points:
\begin{equation}
    \text{Pur}(\mathbf{Q}, \mathbf{Y}) = \frac{1}{N} \sum_{k=1}^K \underset{j = 1 \ldots C}{\max} |Q_k \cap Y_j|.
\end{equation}
High purity is easy to achieve when the number of samples per cluster is small; in particular, purity equals 1 if $K = N$. Thus, purity cannot be used to trade off the validity of the clustering against the number of clusters.

\paragraph{Unsupervised clustering accuracy}

Classification accuracy is the number of samples assigned to the correct class divided by the total number of samples. It can also be used in clustering as an external quality measure if labels are available and if the number of clusters is equal to the number of classes. It consists in the accuracy of the resulting classification using the best one-to-one mapping $m$ between clusters and class labels: 
\begin{equation}
    \text{Acc}(\mathbf{Q}, \mathbf{Y}) = \frac{1}{N} \underset{m}\max \sum_{k=1}^K |Q_k \cap Y_{m(k)}|.
\end{equation}
The best mapping can be found using the Hungarian assignment algorithm, also known as the Kuhn-Munkres algorithm. One minus the accuracy is also known under the names \textit{minimum matching distance} (MMD) or \textit{0-1 loss}. As it requires to have $K = C$, it is generally necessary to cluster the map prototypes into fewer classes \cite{Vesanto2000,Azzag2008}. 

\paragraph{Class scatter index}

The class scatter index (CSI), introduced in \cite{Elend2019}, is an external SOM quality index that measures the scattering of ground-truth class labels on the map. It is based on the idea that on a \textit{good} map, classes should be distributed into few distinct groups of neighboring units, and not scattered all over the map. This is important for interpretation, because it allows to associate map areas with particular classes. In practice, it computes the number of \textit{groups of neurons} on the map, where for a given class, a group is defined as a set of neighboring neurons that have at least one sample belonging to that class assigned to them. Finally, CSI is the average number of groups over all classes.

\section{SOMperf: module description and usage examples}

SOMperf is a Python module gathering performance metrics for self-organizing maps. It depends on the libraries \texttt{numpy}, \texttt{scipy}, \texttt{pandas} and \texttt{scikit-learn}. This section describes the structure of the module and provides several usage examples. SOMperf is divided into following sub-modules:
\begin{itemize}
    \item \texttt{metrics}: performance metric functions.
    \begin{itemize}
        \item \texttt{external}: external indices.
        \item \texttt{internal}: internal indices.
    \end{itemize}
    \item \texttt{utils}: utility functions.
    \begin{itemize}
        \item \texttt{neighborhood}: neighborhood kernel functions (gaussian, window, etc.). Only used in \texttt{distortion} for now.
        \item \texttt{topology}: distance functions on grid topologies (rectangular, hexagonal, etc.).
    \end{itemize}
\end{itemize}

Metric functions usually take as arguments a matrix containing the SOM code vectors, the data matrix and/or the distance matrix, to avoid recomputing all  pairwise distances. In addition, they need a distance function on the map. A code example for computing topographic error with a rectangular map topology is given below:

\begin{Verbatim}[commandchars=\\\{\}]
    \PYG{k+kn}{from} \PYG{n+nn}{somperf.metrics} \PYG{k+kn}{import} \PYG{n}{topographic\PYGZus{}error}
    \PYG{k+kn}{from} \PYG{n+nn}{somperf.utils} \PYG{k+kn}{import} \PYG{n}{rectangular\PYGZus{}topology\PYGZus{}dist}
    \PYG{n}{map\PYGZus{}size} \PYG{o}{=} \PYG{p}{(}\PYG{l+m+mi}{10}\PYG{p}{,} \PYG{l+m+mi}{10}\PYG{p}{)}
    \PYG{n}{dist\PYGZus{}fun} \PYG{o}{=} \PYG{n}{rectangular\PYGZus{}topology\PYGZus{}dist}\PYG{p}{(}\PYG{n}{map\PYGZus{}size}\PYG{p}{)}
    \PYG{n}{x} \PYG{o}{=} \PYG{o}{...} \PYG{c+c1}{\PYGZsh{} data matrix}
    \PYG{n}{som} \PYG{o}{=} \PYG{o}{...} \PYG{c+c1}{\PYGZsh{} SOM code vectors matrix}
    \PYG{n}{te} \PYG{o}{=} \PYG{n}{topographic\PYGZus{}error}\PYG{p}{(}\PYG{n}{dist\PYGZus{}fun}\PYG{p}{,} \PYG{n}{som}\PYG{p}{,} \PYG{n}{x}\PYG{p}{)}
\end{Verbatim}

\begin{figure}
    \centering
    \includegraphics[width=0.30\textwidth]{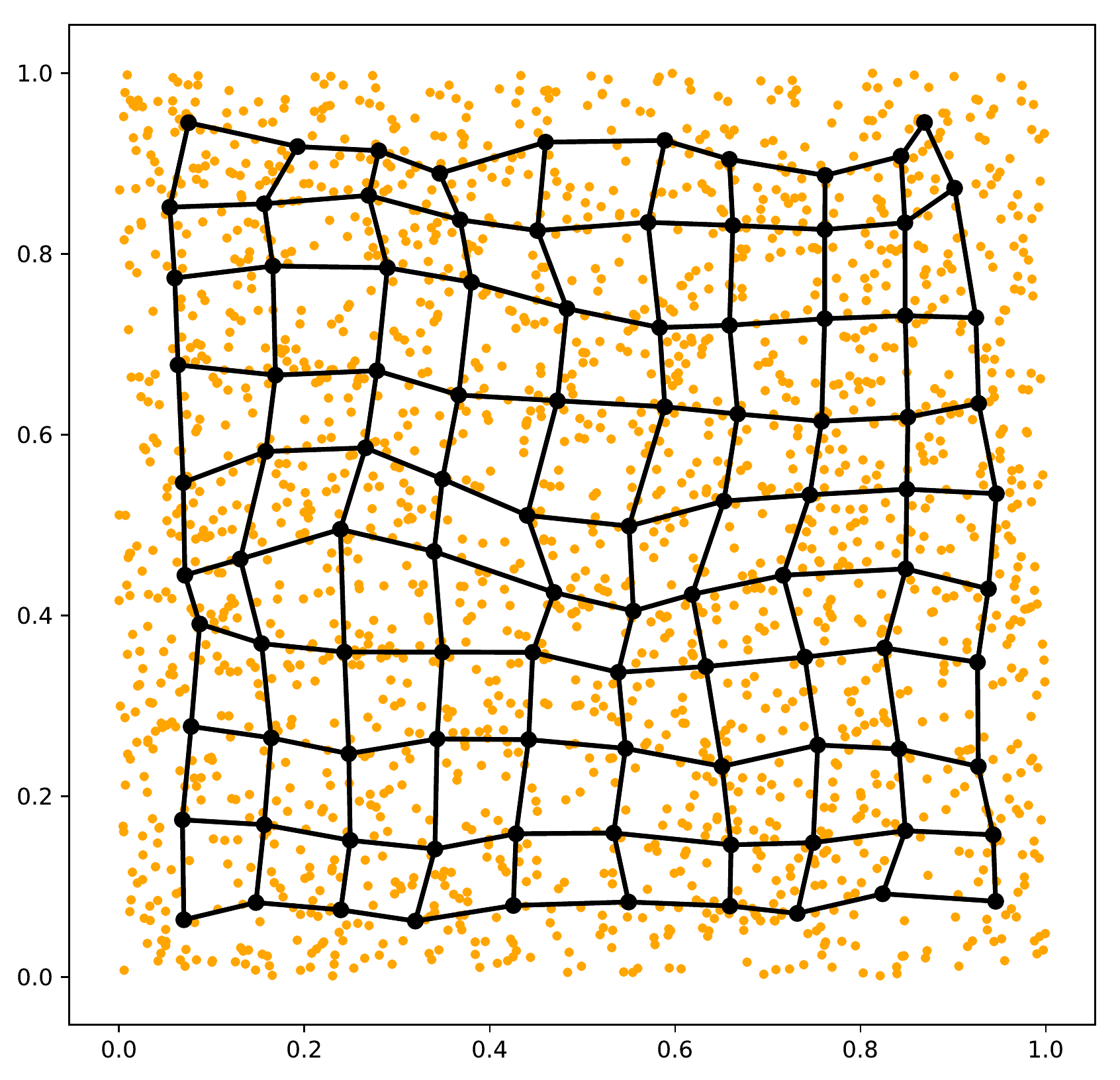}
    \includegraphics[width=0.30\textwidth]{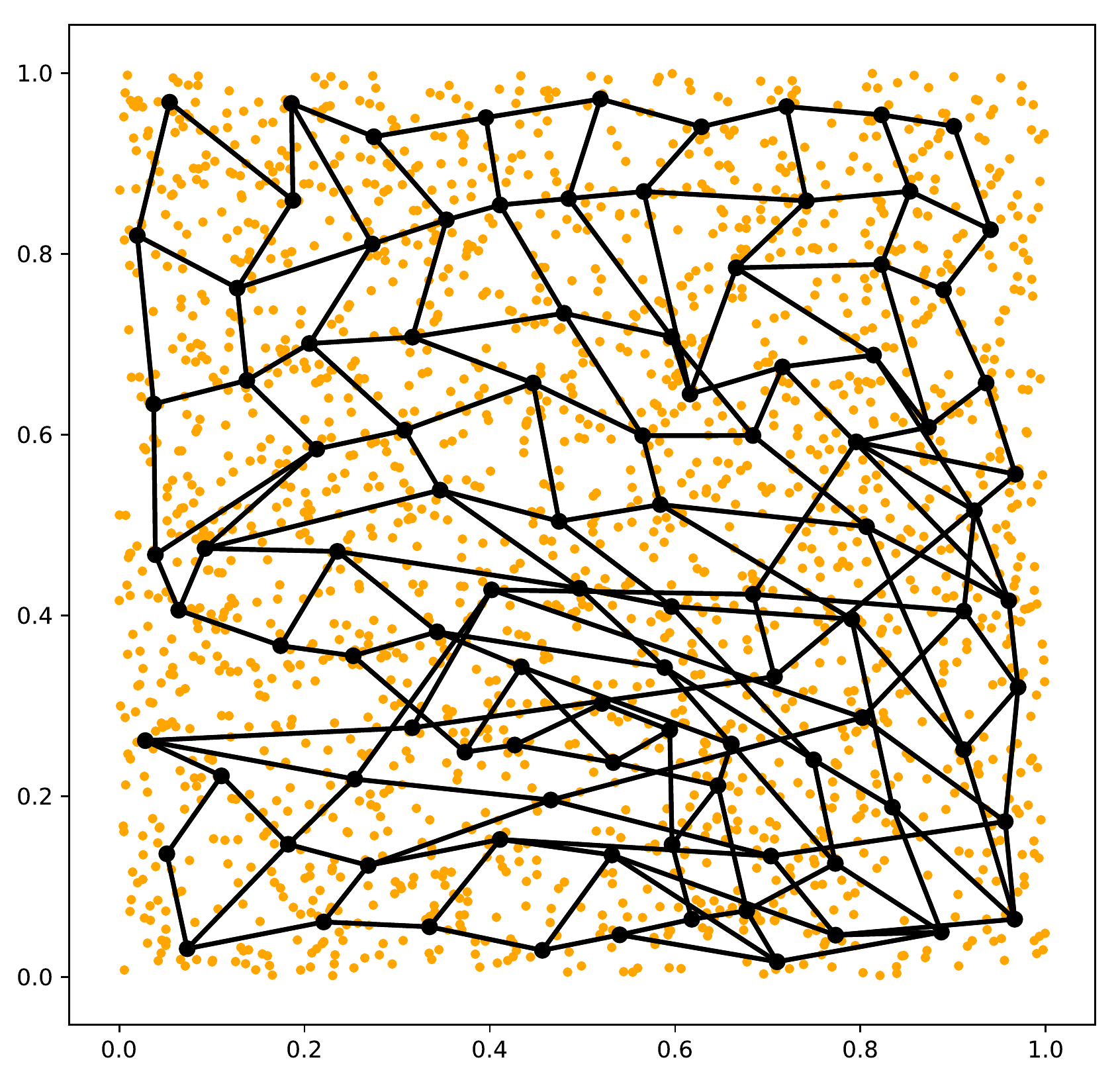}
    \includegraphics[width=0.30\textwidth]{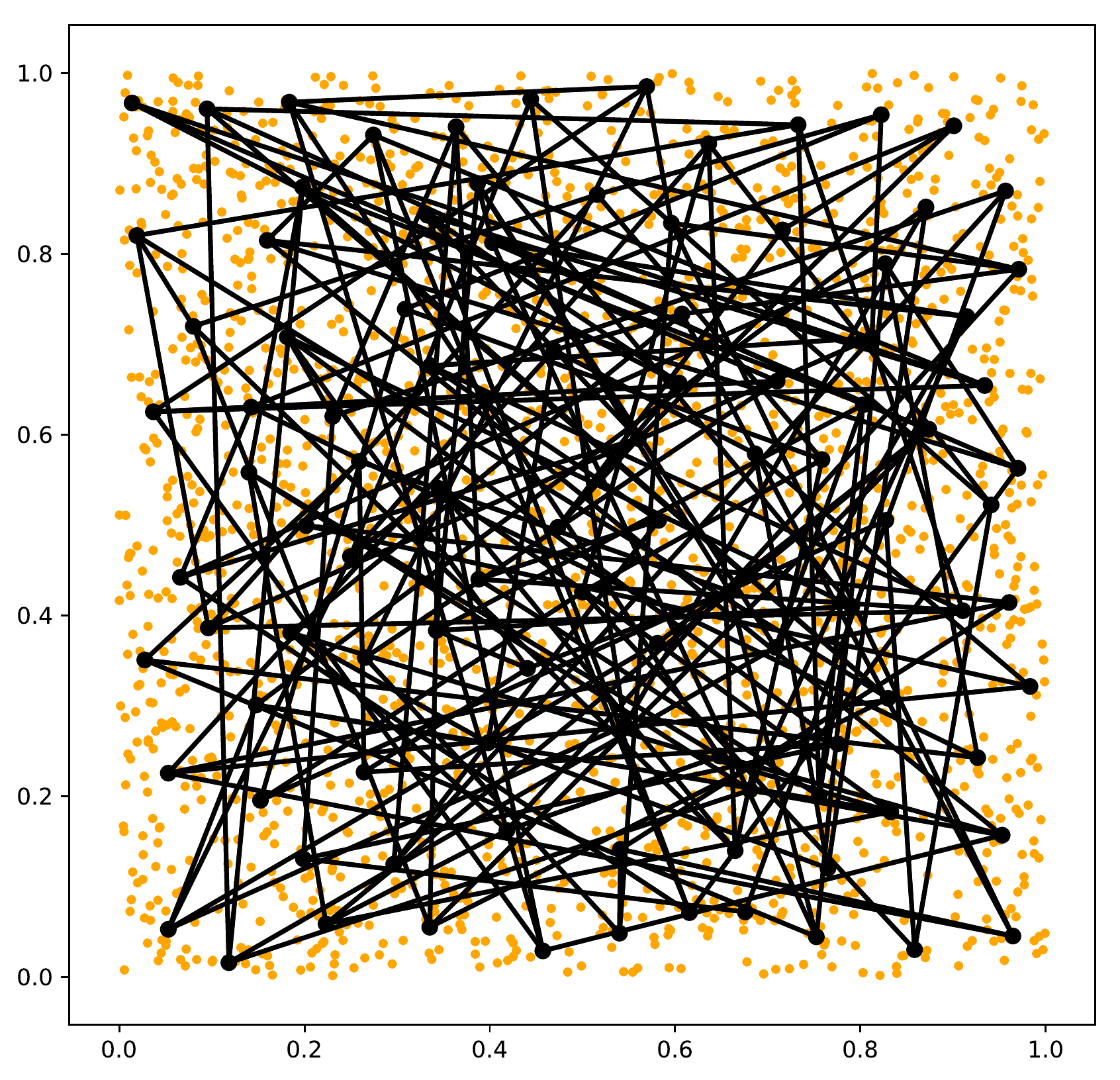}\\
    \includegraphics[width=0.22\textwidth]{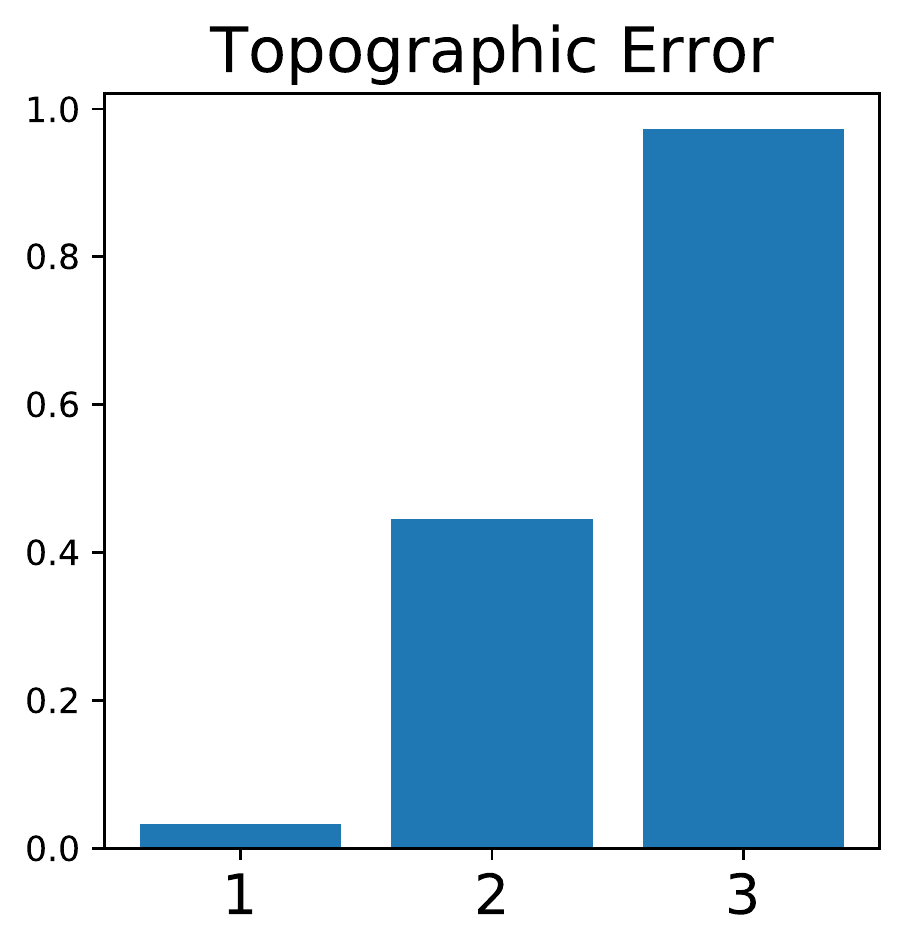}
    \includegraphics[width=0.22\textwidth]{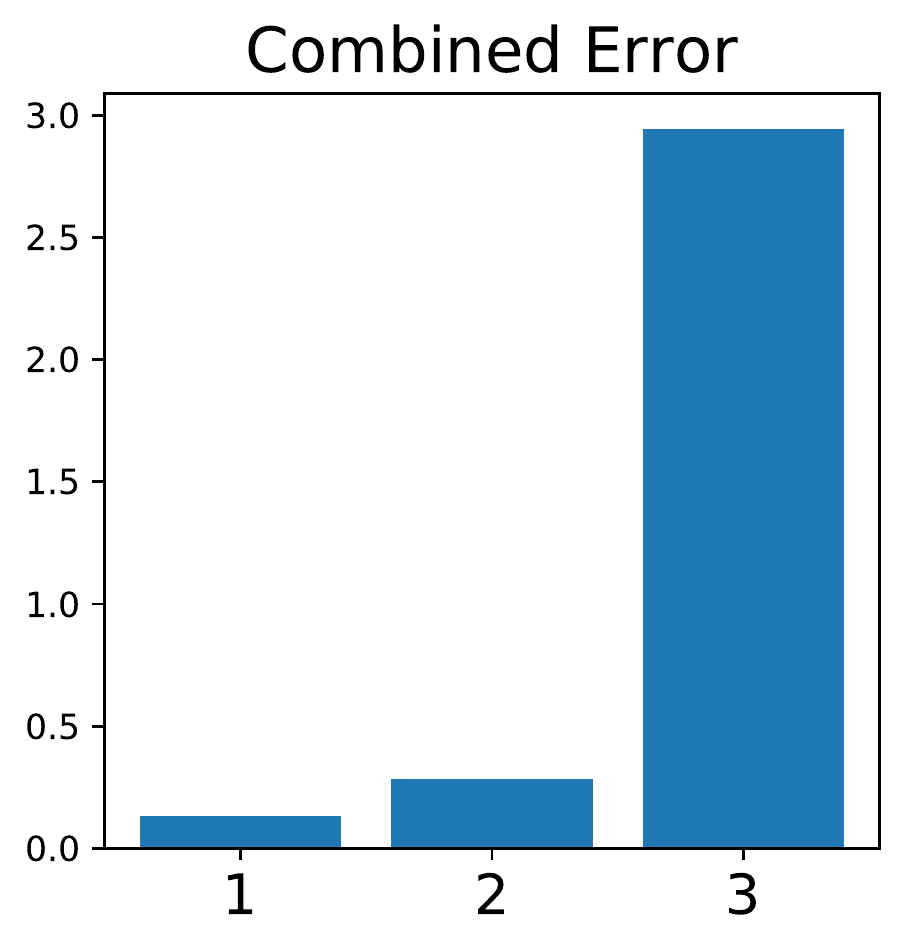}
    \includegraphics[width=0.22\textwidth]{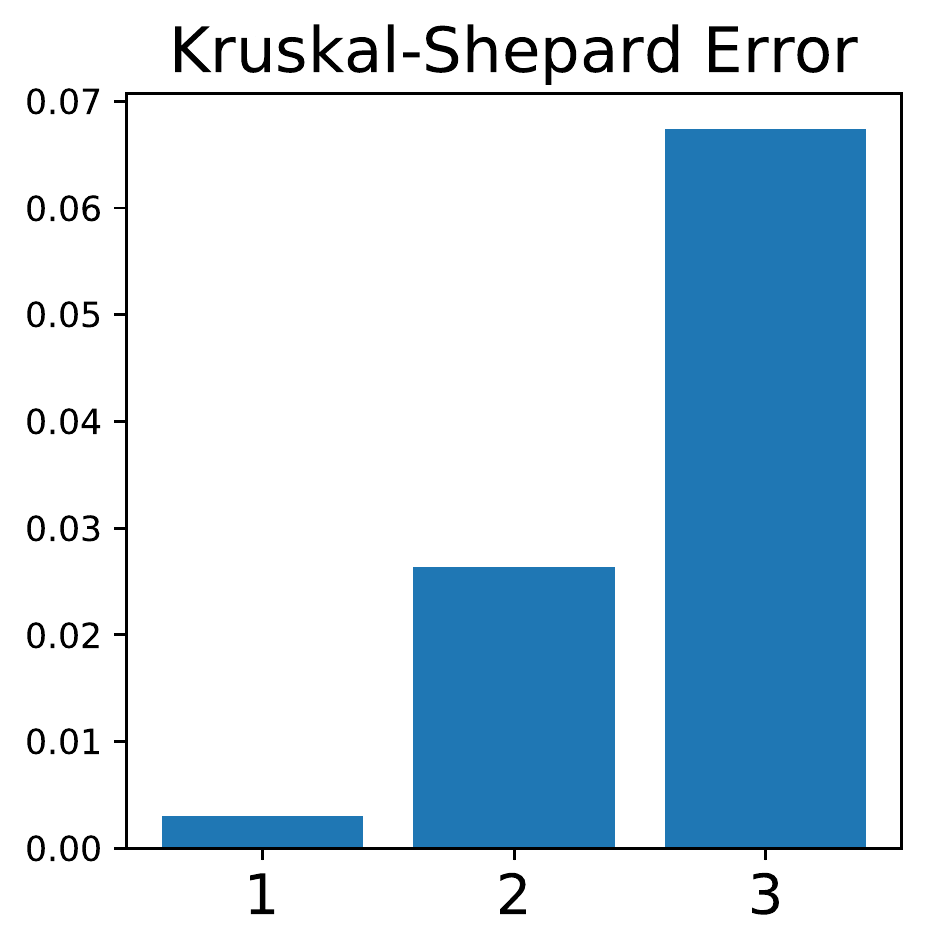}
    \includegraphics[width=0.22\textwidth]{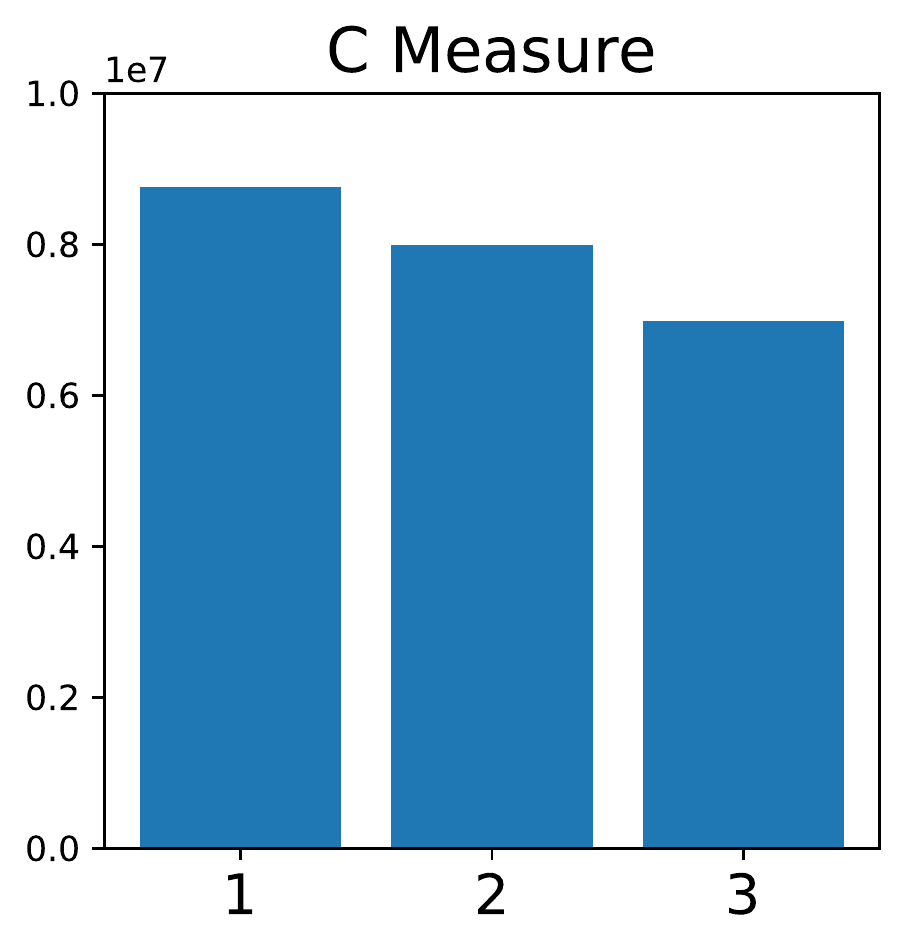}
    \caption{SOM maps representing a uniform distribution with three different levels of topographic organization. Topographic, combined and Kruskal-Shepard errors and C measure behave as expected as a function of organization.}
    \label{fig:perf-examples}
\end{figure}

Figure~\ref{fig:perf-examples} shows three SOM maps trained on a square uniform distribution, with three different levels of topographic organization. Topographic, combined and Kruskal-Shepard error increase as the map becomes more disordered, and the C measure decreases.
\begin{figure}
    \centering
    \includegraphics[width=0.37\textwidth]{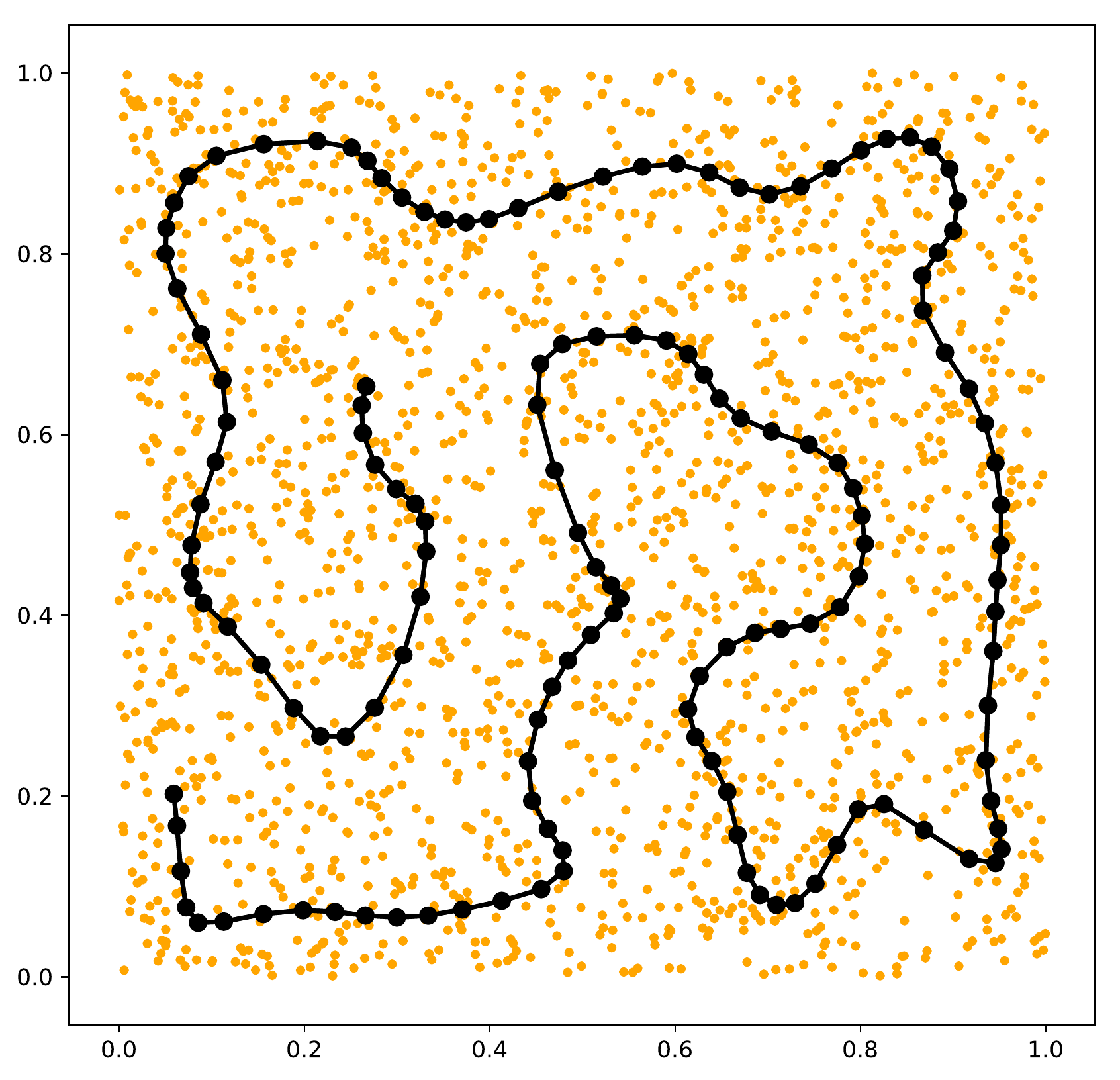}
    \includegraphics[width=0.48\textwidth]{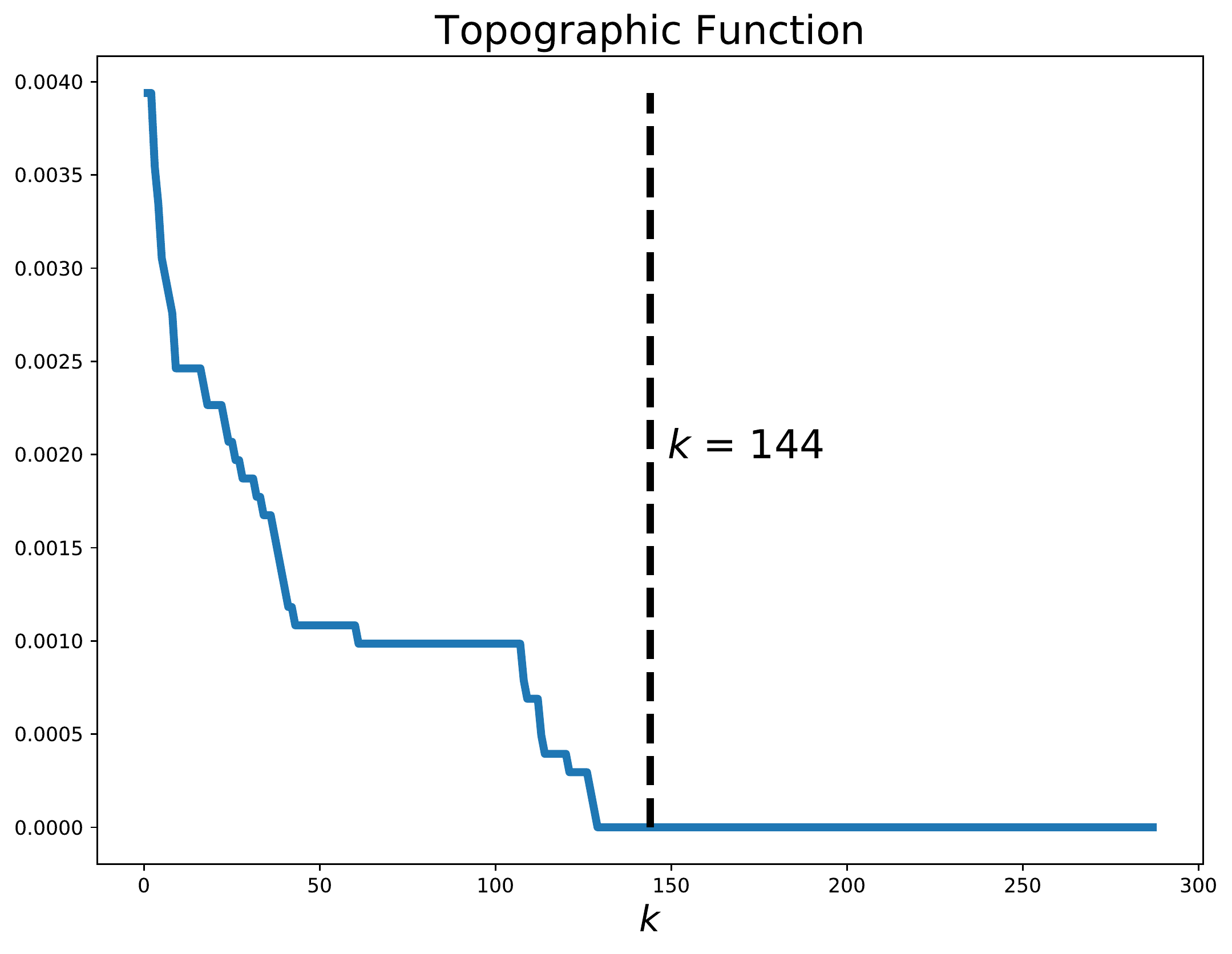}
    \caption{Reproduction of the example in \cite{Villmann1994}. The topographic function vanishes when $k$ approaches the length of the 1D map.}
    \label{fig:tf-examples}
\end{figure}
\begin{figure}
    \centering
    \includegraphics[width=0.15\textwidth]{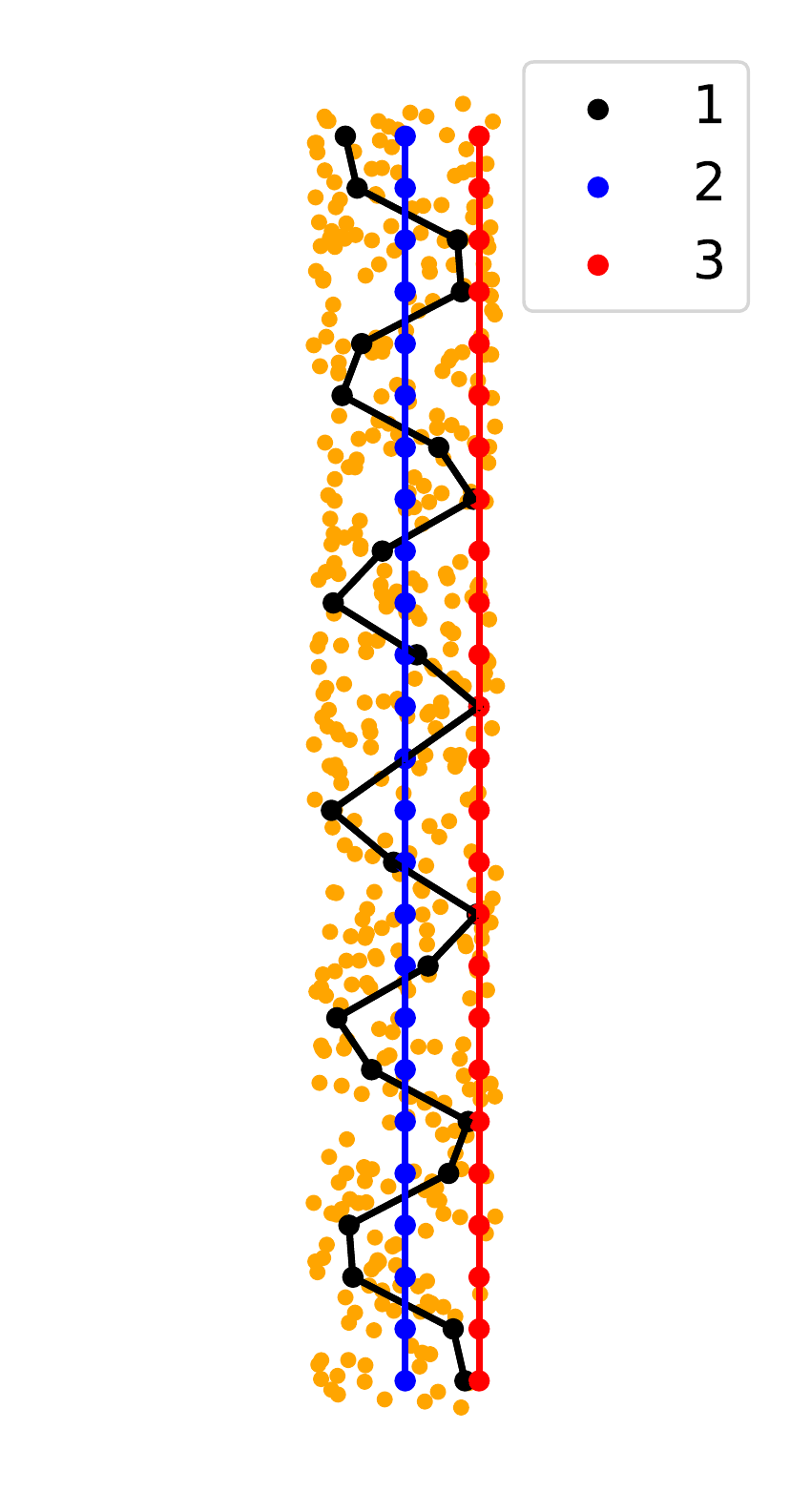}
    \includegraphics[width=0.82\textwidth]{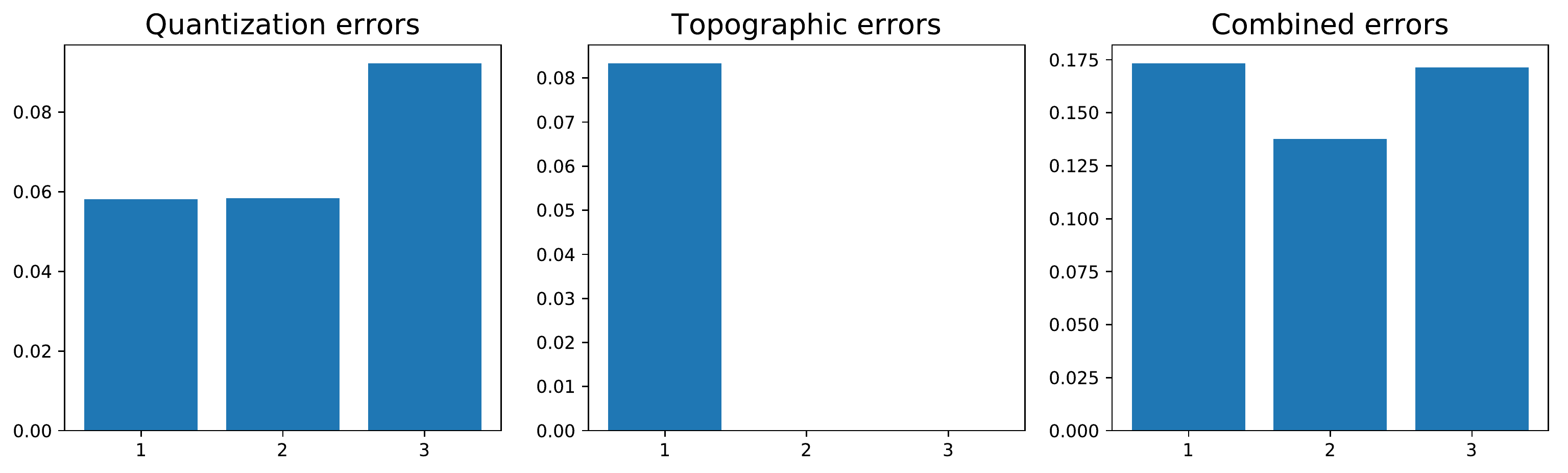}
    \caption{Reproduction of the example in \cite{Kaski1996} with a stripe distribution and 1D SOM maps with three different configurations. Among quantization error, topographic error and combined error, only the latter is able to indicate the best solution, i.e. the straight line.}
    \label{fig:snake-examples}
\end{figure}
Figure~\ref{fig:tf-examples} reproduces the experiments described in \cite{Villmann1994}: the topographic function vanishes when its argument $k$ approaches the length of a 1-dimensional map. On figure~\ref{fig:snake-examples}, we reproduced the example from \cite{Kaski1996}, with three different solutions of a 1-dimensional SOM trained on a 2D rectangular stripe. SOlution (1) approximates well the data (low quantization error), but is too \emph{complex} and less smooth (high topographic error), whereas solution (3) has a bias (high QE) but is a simple straight line (low TE). Only combined error is able to indicate the best comprise, i.e. solution (2). A large number of additional experiments are available online as a notebook\footnote{\url{https://github.com/FlorentF9/SOMperf/blob/master/tests/SOMperf-Tests.ipynb}}, including examples from the original papers.

\section{Conclusion}

In this paper, we reviewed various internal and external performance metrics for SOM and introduced the SOMperf Python module, enabling practitioners to easily evaluate their models. Future work perspectives include the computation of per-unit metrics, a SOM visualization module, as well as distance functions between self-organized models. In addition, other more recent SOM quality metrics could be implemented, such as the map embedding accuracy \cite{Hamel2016}.

\clearpage
\bibliographystyle{unsrt}
\bibliography{references}

\end{document}